\documentclass[letterpaper, 10 pt, conference]{ieeeconf}  
\IEEEoverridecommandlockouts                              
\usepackage[noadjust,sort]{cite}
\usepackage{amsmath,amssymb,amsfonts,graphicx,textcomp,xcolor,multirow,subcaption,tabu,booktabs}
\usepackage{algorithmic}
\usepackage[ruled,vlined]{algorithm2e}
\usepackage[normalem]{ulem}
\usepackage[pagebackref=true,breaklinks=true,colorlinks=true,bookmarks=false,citecolor=blue]{hyperref}
\hypersetup{
colorlinks=true,
linkcolor=blue,
filecolor=magenta,
urlcolor=blue,
citecolor=blue
}

\useunder{\uline}{\ul}{}
\title{\LARGE \bf Auto-Tuned Sim-to-Real Transfer}
\author{Yuqing Du*\thanks{*Equal contribution. Correspondence to \texttt{yuqing\_du@berkeley.edu} and \texttt{oliviawatkins@berkeley.edu}.}$^{1}$, Olivia Watkins*$^{1}$, Trevor Darrell$^{1}$, Pieter Abbeel$^{1}$, Deepak Pathak$^{2}$\\
$^{1}$ UC Berkeley, $^{2}$ Carnegie Mellon University
}
\begin{document}
\maketitle
\thispagestyle{empty}
\pagestyle{empty}
\begin{abstract}
Policies trained in simulation often fail when transferred to the real world due to the `reality gap' where the simulator is unable to accurately capture the dynamics and visual properties of the real world. Current approaches to tackle this problem, such as domain randomization, require prior knowledge and engineering to determine how much to randomize system parameters in order to learn a policy that is robust to sim-to-real transfer while also not being too conservative. We propose a method for automatically tuning simulator system parameters to match the real world using only raw RGB images of the real world without the need to define rewards or estimate state. Our key insight is to reframe the auto-tuning of parameters as a search problem where we iteratively shift the simulation system parameters to approach the real world system parameters. We propose a Search Param Model (SPM) that, given a sequence of observations and actions and a set of system parameters, predicts whether the given parameters are higher or lower than the true parameters used to generate the observations. We evaluate our method on multiple robotic control tasks in both sim-to-sim and sim-to-real transfer, demonstrating significant improvement over naive domain randomization. Project videos at \url{https://yuqingd.github.io/autotuned-sim2real/} and code available at \url{https://github.com/yuqingd/sim2real2sim_rad}.
\end{abstract}

\section{Introduction}
Recently, there has been encouraging progress in deep learning techniques for learning complex control tasks ranging from locomotion~\cite{tan2018simtoreal,lee2020learning} to manipulation~\cite{openai2018learning}. Many of these approaches, especially deep reinforcement learning, rely on learning in simulation before transferring to real world. However, this transfer is often difficult due to the `reality gap'~\cite{Jacobi:1995:NRG:645300.648380}, which arises from the difficulty of accurately simulating the dynamics and visuals of the real world. One way to circumvent the gap is to learn directly from data collected by real robots \cite{DBLP:journals/corr/LevinePKQ16}, but this process is often time-consuming, potentially unsafe, and challenging when a reward function for the real world task is unknown.

\begin{figure}[t]
 \centering
 \includegraphics[scale=.15]{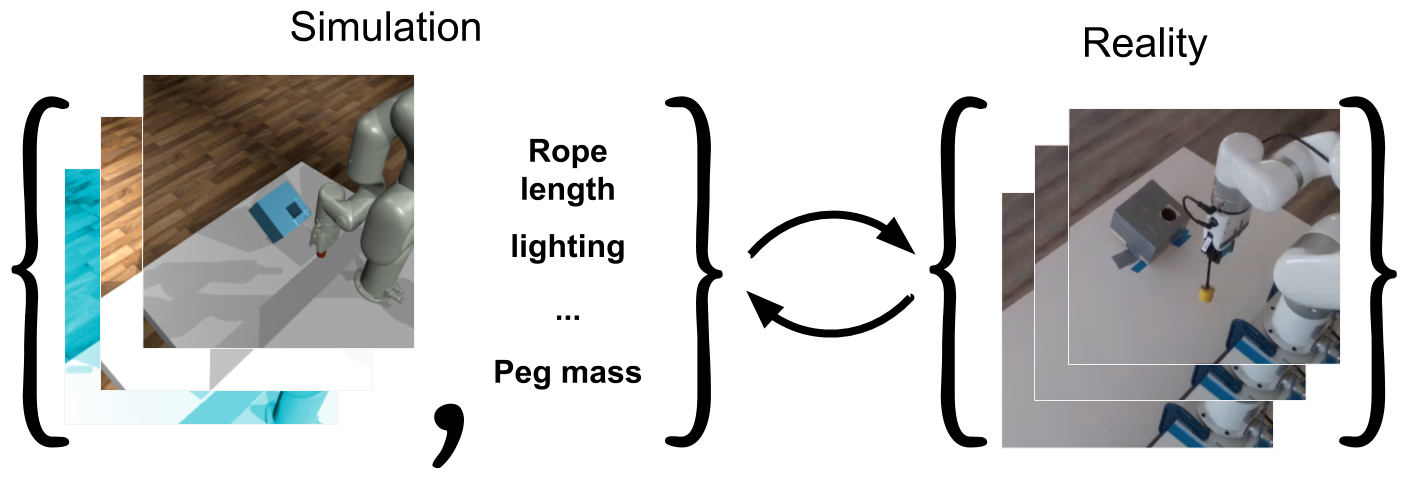}
 \caption{Policy is trained on domain-randomized data in simulation and transferred to the real-world. Instead of relying on manual engineering to tune the simulation, we automatically learn to adjust the system parameters of the simulator to generate trajectories that most closely match reality by only using raw observation images from the real world without any instrumentation or any knowledge of state and reward.}
 \vspace{-0.1in}
 \label{fig:teaser}
\end{figure}

On the other hand, training with simulators provides vast amounts of varied data quickly while being safer and less costly. Yet transferring a policy learned in simulation to the real world often fails due to small simulation discrepancies which errors that compound over time in the real world.
Current techniques that improve transfer success rates can be categorized into three kinds: system identification, domain adaptation, and domain randomization. These methods have certain tradeoffs: (1) Traditional system identification~\cite{Ljung:1986:SIT:21413} involves defining and tuning parameters that describe the real world system accurately, but often takes a lot of time and leads to systems that have to be re-tuned over time. (2) Domain adaptation methods involve learning a discriminator or other type of predictor which incentivizes a model to learn features invariant to the shift between training (simulation) and test (reality) distributions \cite{ganin2015domainadversarial}, however it often requires vast amounts of real world image data. In addition, there are some aspects of the domain shift the agent should not be invariant to (e.g. a robot may need to adjust its policy if the real world has much higher friction than the simulation). (3) Domain randomization methods involve training the model across a variety of simulated environments with randomized visuals \cite{tobin2017domain} or dynamics \cite{peng2018sim} with the hypothesis that given enough variability in the simulations, the real world can be viewed as another `randomized' instance, thus allowing the model to generalize to the real world.

This paper belongs to the broad category of methods that employ domain randomization. While this technique has become widely used in robotics, it requires task-specific expert knowledge to correctly determine which parameters to randomize and to what extent they should vary in order to learn policies that are robust to transfer without being too conservative. Domain randomization can also potentially make the task too difficult if too much randomization impedes the model's ability to learn a successful policy \cite{DBLP:journals/corr/abs-1806-07851}. Recent works \cite{simopt,chang2020sim2real2sim} attempt to close the gap between simulation and real-world via trajectory matching but require instrumentation of the real-world (via DART and AR tags, respectively) to obtain true state, which is often time-consuming and unrealistic.

Thus we ask the question: can we auto-tune the system parameters of simulation to match that of reality by just using raw observations (e.g. images) of the real world, thus bypassing the need for instrumenting the real world?  We propose a task-agnostic reinforcement learning (RL) method that aims to reduce the amount of task-specific engineering required for domain randomization of both visual and dynamics parameters. Our method uses a small amount of real world video data and learns to slowly shift the simulation system parameters to become more similar to the real world. As directly predicting these parameters from observations is challenging, our key insight is to instead predict whether the current parameters are higher, lower, or close to the real world values in a given sequence of observations. Thus, we are able to leverage both the benefits of simulation: an abundance of varied data and ease of using a shaped reward function, as well as the benefits of the real world: providing a ground truth for improving our simulation. By moving the simulation data distribution closer to the real world data distribution, the simulation should become more representative of reality, leading to a policy more likely to succeed in the real world.

Our main contributions are:
\begin{itemize}
\item Proposing an automatic system identification procedure with the key insight of reformulating the problem of tuning a simulation as a search problem.
\item Designing a Search Param Model (SPM) that updates the system parameters using raw pixel observations of the real world.
\item Demonstrating that our proposed method outperforms domain randomization on a range of robotic control tasks in both sim-to-sim and sim-to-real transfer.
\end{itemize}

\section{Related Work}
\label{sec:related}

\paragraph{Sim2Real via Domain Randomization}
Domain randomization \cite{tobin2017domain}, the practice of training a model on several variants of a simulated environment, often enables better transfer to the real world.  Prior works have randomized visuals, \cite{pinto2017asymmetric}, system dynamics \cite{peng2018sim}, and the placement of objects \cite{CAD2RL}.  Typically, a human must use heuristics, domain knowledge, and trial-and-error to hand-engineer the distribution of simulation parameters used for training. Domain randomization alone is often insufficient for tasks which require precise motions or dexterous manipulation.  \cite{adr} address this through automatic domain randomization, where they begin with a single simulation variant and then automatically increase parameter ranges as the agent learns.  The wide final distribution of parameters trained on may not reflect the real-world distribution of environments.  In contrast, our work incorporates real data in order to learn a distribution of simulation parameters which matches the real world. \cite{RoboImitationPeng20} learns mobile robot policies in simulation which condition on an encoded version of the simulation parameters of the environment. In the real world, they search for a set of simulation parameters which enable the policy to achieve high reward in the real world. While we also aim to find the ``simulation parameters'' of the real world, we do this without requiring any real-world reward supervision. \cite{yu2017preparing} similarly learns a policy conditioned on a set of system parameters, but they obtain system parameters for the test environment by learning a model which predicts system parameters from observations.  However, this work only demonstrates sim-to-sim transfer on state-based observations, a much easier problem than the image-based sim-to-real tasks we tackle.

\paragraph{Sim2Real with Real Data}
While training only in the real world circumvents the reality gap, work in this area is often impractical and time consuming. The recent SimOpt work \cite{simopt} is most closely related to our work. Here, partial observations of the real world are leveraged to better inform simulation randomization such that the performance of policies in simulation more closely matches their performance in the real world, while avoiding reward instrumentation in the real world. However, SimOpt relies on continuous object tracking in the real world in order to make trajectory comparisons between simulation and the real world, whereas our system solely operates through pixel-based observations. \cite{zhu2017fast} similarly uses real-world observations to perform system identification, but it requires real-world state and reward estimation. \cite{jeong2019selfsupervised} tackles the problem of learning robotic manipulation from pixels through sequence-based self-supervision. They execute the learned policy on the real robot to gather sequences of unlabeled real world images which are used with sequences of simulated images to perform domain adaptation. While our approach also uses unlabeled real world images for sim-to-real transfer, we focus on improving our simulation to better match reality.

\section{Background and Preliminaries}
Consider a real-world RL problem defined by a POMDP $\mathcal{M} = (S, O, A, R, \gamma, \xi_{real})$, where $S$ is the unobserved state space, $O$ is the agent's observation space, $A$ is the agent's action space, $R: S \times A \rightarrow \mathbb{R}$ is the reward function, $\gamma$ is the discount factor, and $\xi_{real}$ defines the system parameters of the real world.  These parameters affect both the transition dynamics $P(s_{t+1} | s_t, a_t)$ and the mapping from states to observations $P(o_t|s_t)$. As the reward function depends on the unobserved state $s$, in practice it is challenging to fully instrument the real world in order to give this reward to the agent. Instead, we train the agent in a simulation where $S$ and $R$ are easily accessible. The dynamics and visuals of the simulator are defined by simulator system parameters, $\xi_{sim}$.

\begin{figure*}[t]
 \centering
 \includegraphics[width=0.88\linewidth]{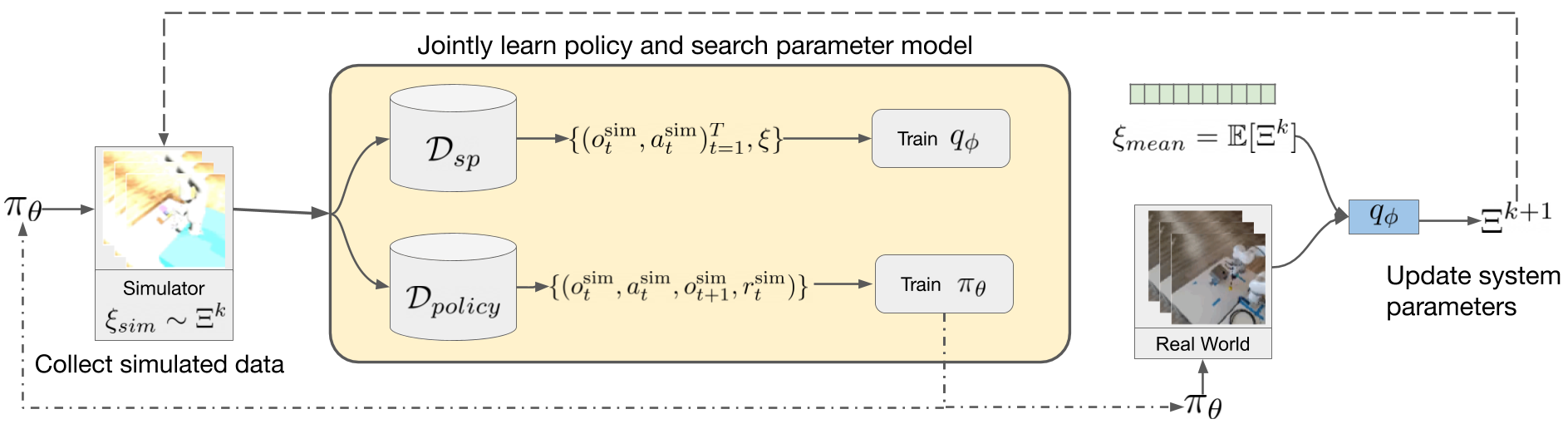}
 \vspace{-0.05in}
 \caption{Overall System: Using any off-the-shelf RL algorithm, we use simulated data to train both our policy and a Search Parameter Model (SPM) $q_{\phi}$ which predicts whether a candidate set of system parameters $\xi$ is higher or lower than those which produced an observed trajectory. We iteratively update our simulation by running our policy in the real world and using our SPM to predict which direction to update our simulator to make it closer to the real world.}
 \vspace{-0.05in}
 \label{fig:sys_diagram}
\end{figure*}

Domain randomization samples the simulator parameters from a distribution of $N$ system parameters, $\xi_{sim} \sim \Xi \subset \mathbb{R}^N$. In our work, $\Xi$ is a uniform distribution with mean $\xi_{mean}$ and range proportional to the mean, scaled by a fixed hyperparameter. Each time the simulator resets, the simulator is reparametrized with new sampled parameters $\xi_{sim}$. The goal of domain randomization is to train the policy in simulation to maximize $\mathbb{E}_{\xi_{sim} \sim \Xi} [R(\tau_{\xi_{sim}})]$, where $\tau_\xi$ is a trajectory collected in a simulator parametrized with $\xi$. This policy can be trained with any reinforcement learning algorithm and in this work we use SAC \cite{haarnoja2018soft}. By training on an appropriate distribution of simulation parameters, the policy will hopefully generalize to the real world.

\section{Our Approach: Auto-tuning Sim2Real}
A challenge of domain randomization is if $\Xi$ does not cover $\xi_{real}$, then it is unlikely that a policy trained on environments randomized by $\Xi$ will transfer well to the real world. Therefore, it is common practice to choose a wide enough distribution  However, this introduces an additional challenge of training a single universal policy that optimizes expected reward across all $\xi \sim \Xi$, which can lead to an overly conservative policy and may be computationally expensive. Thus in order to achieve good performance with domain randomization, we must define $\Xi$ using parameters which are close enough to $\xi_{real}$ that we can train on a narrow distribution of parameters.

Standard practice is to use expert knowledge and spend time manually engineering the environment such that parameters $\xi_{mean}$ reasonably approximate $\xi_{real}$. Prior work \cite{simopt} has proposed selecting $\xi_{real}$ by comparing trajectories from differently parametrized simulations and from the real world, choosing the simulation parameters which produce trajectories most similar to the real world. However, measuring trajectories requires obtaining state-space information in the real world. This is often impractical in practice, so we propose an approach that only uses raw pixel observations in the real world to find $\xi_{mean}$ and auto-tune our simulator.

\subsection{Reformulating Auto-tuning as a Search Problem}
To overcome the challenges of domain randomization and avoid determining the full state in the real world, we propose an approach to automatically find $\xi_{mean} \approx \xi_{real}$ using a function $f : (o_{1:T}, a_{1:T}) \rightarrow \xi$ that maps a sequence of observations and actions to their corresponding system parameters. In fact, this is the exact problem studied in system identification. Learning such a function is challenging for two reasons. Firstly, performing system identification for complex dynamical systems from high-dimensional inputs, such as images, remains challenging. Consider the example of a robot opening a cabinet. It is challenging to identify the exact values of the damping on the cabinet hinge, the friction of the handle against the robot's arm, and so forth, solely from observations. Secondly, if the true parameters are outside of the training distribution of simulator parameters, predicting the parameters directly will likely not work. We propose to circumvent these issues by \textit{reformulating the auto-tuning problem as a search procedure}. Rather than predicting $\xi_{real}$ exactly, it is much easier to train a model to detect which half-space $\xi_{real}$ resides in.

We propose a Search Param Model (SPM) which is a binary classifier $q_{\phi}: (o_{1:T}, a_{1:T}, \xi_{pred}) \rightarrow (0, 1)^{N}$ which predicts the probability that the true set of system parameters which generated the trajectory $o_{1:T}$ is higher, lower, or about equal to the given $\xi_{pred}$ for each of the $N$ system parameters. We use this binary classifier to iteratively `auto-tune' $\xi_{mean}$ to be closer to $\xi_{real}$.  To do this, we first train the SPM on simulated trajectories $(o_{1:T}, a_{1:T})$ generated using the current simulation system parameters $\xi_{sim}$.  We then randomly sample candidate system parameters $\xi_{pred}$ and compute labels $L= \xi_{sim} > \xi_{pred} \in (0, 1)^N$, then train using logistic regression.

We train the SPM in two stages: first pretraining with the initial $\Xi$ using simulated samples from a random policy, then in an iterative fashion jointly with policy training, described in Section \ref{subsec:joint}. After pretraining, we can use the SPM to update $\xi_{mean}$ to be closer to $\xi_{real}$ by occasionally collecting a real world trajectory $(o_{1:T}^{real}, a_{1:T}^{real})$.  Using the SPM, we compute $q_{\phi}(o_{1:T}^{real}, a_{1:T}^{real}, \xi_{mean})$, which predicts whether our current parameters are too high or too low. We update each parameter in $\xi_{mean}$ based on the confidence of the prediction, either increasing or decreasing the parameter value incrementally for confident predictions or maintaining the same parameter for less confident predictions.

\subsection{Jointly Learning to Act and Auto-tune Simulation} \label{subsec:joint}

Our proposed algorithm interleaves the process of learning a policy in simulation with continuously updating the simulation using the SPM. It is often not sufficient to just pretrain the SPM, as $\Xi$ may change significantly from the initial distribution as auto-tuning continues. Therefore we continue to train and update the SPM as we slowly shift $\Xi$. An alternating training process is necessary as observing the effect of certain parameters (eg. contact friction) requires learning a reasonable policy to interact with the environment first. We also want to prevent overfitting the policy to a poorly parametrized simulation if the SPM makes incorrect predictions. The alternating training algorithm is detailed in Algorithm \ref{alg} and our system is illustrated in Figure \ref{fig:sys_diagram}.

This joint training procedure requires carefully generating the simulation data. In order to gain intuition about the effects of different system parameters on the environment, the SPM needs to be trained on a wide distribution of parameters. On the other hand, the agent's policy often struggles to learn when the distribution of simulation parameters is too large. To address this, we use separate buffers to train the SPM and our policy, $\mathcal{D}_{sp}$ and $\mathcal{D}_{policy}$ respectively. The system parameters used to generate trajectories for each dataset are sampled from uniform distributions centered around the same distribution mean $\xi_{mean}$, but with different range hyperparameters $r_{sp}$ and $r_{policy}$ for the SPM and policy buffers respectively, with $r_{sp} > r_{policy}$.

$\mathcal{D}_{sp}$ also stores full sequences of $(o_t, a_t)$ rather than individual transitions, as certain parameters can only be gleaned from seeing a sequence of observations.

\begin{algorithm}[]
\SetAlgoLined
Initialize buffers $\mathcal{D}_{sp}$, $\mathcal{D}_{policy}$, agent, and SPM $q_\phi$\;
Initialize simulator with $\Xi^0$, parametrized by $\xi_{mean}$\;
Fill  $\mathcal{D}_{sp}$ with simulated trajectories from a random policy and pretrain $q_\phi$\;
\For{step k=1:K}{
\texttt{//Policy Training Phase} \\
At each reset, sample sim params $\xi_t^{\text{sim}} \sim \Xi^{k}_{policy}$ \;
\For{training step t=1:policy\_itrs}{
  Collect simulated trajectories and store into $\mathcal{D}_{policy}$\;
Train policy $\pi_\theta$ with data from $\mathcal{D}_{policy}$\;
}
\texttt{//Sim Param Training Phase} \\
At each reset, sample sim params $\xi_t^{\text{sim}} \sim \Xi^{k}_{sp}$ \;
\For{training step t=1:sim\_param\_itrs}{
  Collect simulated trajectories in $\mathcal{D}_{sp}$\;
  Train SPM $q_\phi$ with     $(\{(o_t^{\text{sim}}, a_t^{\text{sim}})\}_{t=1}^T,
  \xi_t^{\text{sim}}) \sim \mathcal{D}_{sp}$\;
}

\texttt{//Update Simulation from Real}\\
Collect samples by running $\pi_\theta$ in the real world\;
Update simulator distribution using the SPM: $\Xi^{k+1} \leftarrow \Xi^{k} + \alpha q_\phi(\{(o^{\text{real}}_t, a^{\text{real}}_t)\}_{t=1}^T, \xi_{mean}^k)$\;
}
\caption{Auto-Tuned Sim2Real}
\label{alg}
\end{algorithm}

\section{Implementation Details}

\paragraph{Policy Learning}
As our method of updating system parameters is independent from the agent policy training, the SPM can be combined with any image-based RL algorithm. In this work we use contrastive learning combined with SAC \cite{laskin_srinivas2020curl} as the image-based RL algorithm.

\paragraph{SPM Learning}
The observations $o_{1:T}$ are encoded with a convolutional encoder which has the same architecture as the policy encoder from \cite{laskin_srinivas2020curl}. We stack 10 frames at a time, but since our trajectories are longer than 10 frames, we use multiple windows and average their predictions. We concatenate the encoded features with the agent's actions and $\xi_{pred}$, which is uniformly sampled from $[0, 2 \times \xi_{mean}$]. As each system parameter can represent very different parameters, we use sinusoidal encoding \cite{mildenhall2020nerf} to encode the real values to the same higher dimensional space. Any state information accessible in both simulation and reality can also be concatenated if desired. Architecture and hyperparameter details can be found in Appendix \ref{appendix:architecture}.

\section{Experiments}\label{sec:experiments}
We explore two questions: a) Can our method update our simulator to the correct system parameters? b) Can our method improve real world return over sim-to-real transfer with naive domain randomization?

\subsection{Sim-to-Sim Transfer}
\begin{figure*}[t]
\includegraphics[width=.33\linewidth]{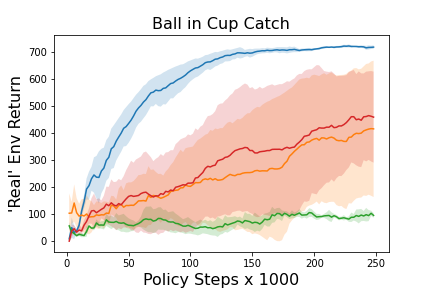}
\includegraphics[width=.33\linewidth]{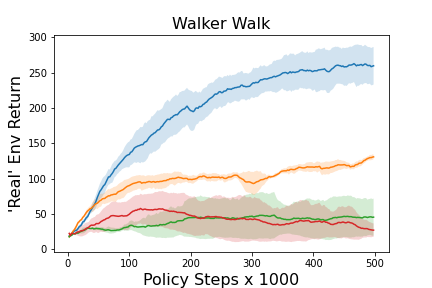}
\includegraphics[width=.33\linewidth]{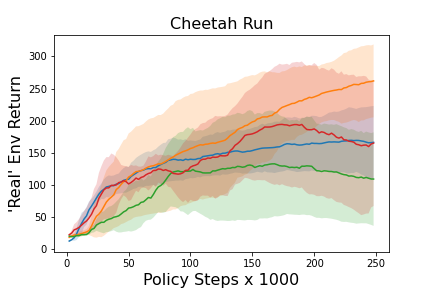}
\includegraphics[width=.33\linewidth]{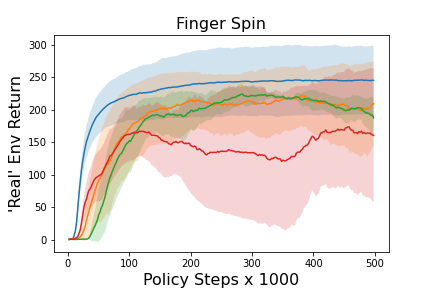}
\includegraphics[width=.33\linewidth]{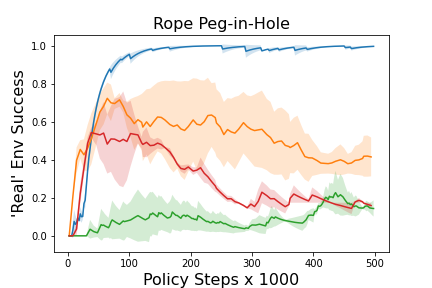}
\includegraphics[width=.33\linewidth]{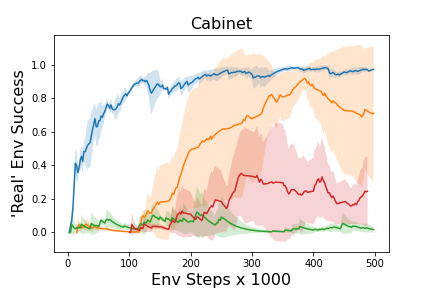}
\begin{center}
\includegraphics[height=.25in]{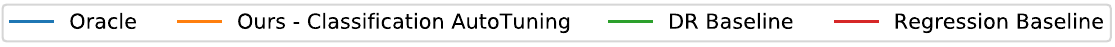}
\end{center}
\vspace{-0.14in}
\caption{Performance on the pseudo-real environment over the course of training. We report reward for the DMC tasks and success on the robotic manipulation tasks.  Different colors represent different experimental conditions: Oracle trains directly on the pseudo-real environment; DR Baseline uses domain randomization; Regression Baseline uses a model to directly predict the system parameters and update it using our method; Our method begins with the same inaccurate mean randomization initialization as the baselines and updates it using our classification method during training.  All runs are averaged over 3 seeds. We see that our auto-tuning method consistently matches or outperforms the baseline.}
\label{fig:sim-return}
\vspace{-0.05in}
\end{figure*}

As proof of concept, we test our method on a suite of 6 sim-to-sim transfer tasks: 4 tasks from the Deepmind Control Suite \cite{deepmindcontrolsuite2018} (Cheetah Run, Walker Walk, Finger Spin, and Ball in Cup Catch) and two robotic arm tasks (Rope Peg-in-Hole and Cabinet Slide).

For each environment we randomize both visual and dynamics parameters as listed below:
\begin{table}[h!]
\vspace{-0.05in}
\centering
\begin{tabular}{l|l} 
 Environment & System Parameters Randomized\\
\midrule
 Ball-in-Cup  &  ball mass; RGB of cup, ball, background \\
 Walker Walk & agent mass; RGB of agent, background\\
 Cheetah Run & agent mass; RGB of agent, background\\
 Finger Spin & mass; damping; RGB of agent, spinner, background\\
  Peg-in-Hole & rope length; peg mass; lighting; RGB of peg, box \\
 Open Cabinet & cabinet friction; lighting; RGB of cabinet, table
\end{tabular}
\label{tab:envs}
\vspace{-0.1in}
\end{table}

For all experiments, we select one set of system parameters as the pseudo-real world and initialize our training distribution such that each parameter was either 2x or 0.5x the value of the pseudo-real world, corresponding to a poorly parametrized simulation.  For our method found the best result with $r_{sp} = 1$ and chose $r_{policy} = .1$. We similarly tuned the domain randomization baseline with a randomization range $r_{dr} \in \{.5, 1\}$ and found the best result with $r_{dr} = .5$. For context, we also train an oracle directly in the pseudo-real world. Figure \ref{fig:sim-return} presents performance for each task when evaluated in the `pseudo-real' world. In all environments, our method matches or exceeds the two baselines: domain randomization and a variant of our method that directly regresses on the simulation parameter values. This suggests that naively initializing domain randomization can make transferring to the real task challenging, and that learning to adjust system parameters by directly regressing on the parameter values is also unlikely to succeed. On the other hand, for many environments our classification method successfully adjusts our simulation parameters to center closely over the correct value.

\begin{figure}[]
\centering
\begin{subfigure}[t]{0.22\linewidth}
    \centering
    \includegraphics[height=.6in]{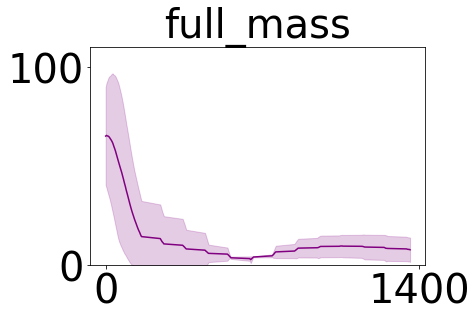}
\end{subfigure}%
~
   \begin{subfigure}[t]{0.22\linewidth}
    \centering
    \includegraphics[height=.6in]{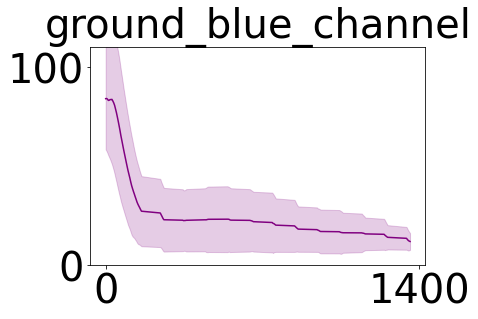}
\end{subfigure}%
~
\begin{subfigure}[t]{0.22\linewidth}
    \centering
    \includegraphics[height=.6in]{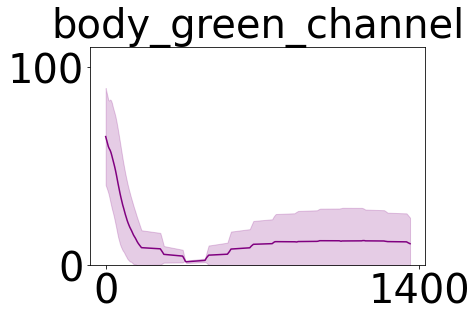}
\end{subfigure}%
~
\begin{subfigure}[t]{0.22\linewidth}
    \centering
    \includegraphics[height=.6in]{images/plots/SPE_walker_body_blue.png}
\end{subfigure}%
\\
    \begin{subfigure}[t]{0.22\linewidth}
    \centering
    \includegraphics[height=.6in]{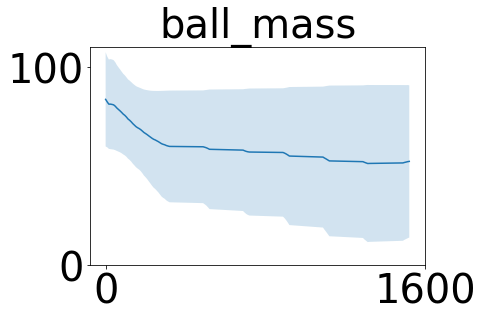}
\end{subfigure}%
~
      \begin{subfigure}[t]{0.22\linewidth}
    \centering
    \includegraphics[height=.6in]{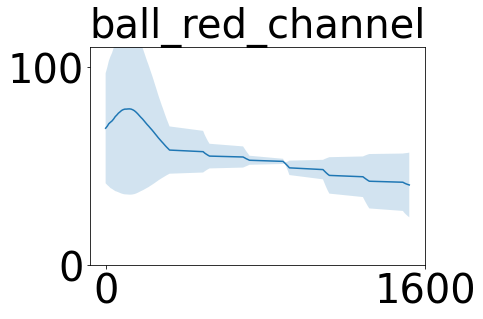}
\end{subfigure}%
~
\begin{subfigure}[t]{0.22\linewidth}
    \centering
    \includegraphics[height=.6in]{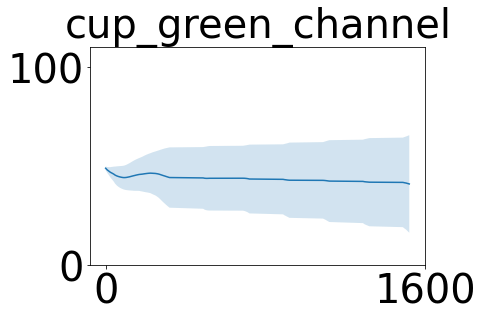}
\end{subfigure}%
~
\begin{subfigure}[t]{0.22\linewidth}
    \centering
    \includegraphics[height=.6in]{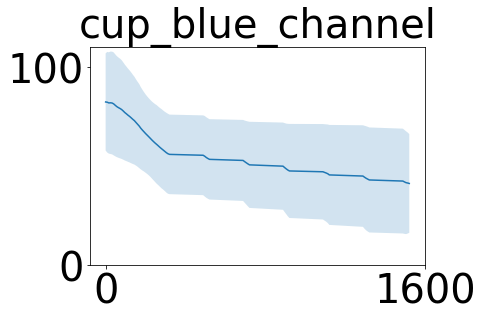}
\end{subfigure}%
\caption{Percent error in our system parameter mean $\xi_{mean}$ over the course of training.  The x-axis displays environment steps ($\times$1000), and the y-axis shows percent error ($100 * |\xi_{mean} - \xi_{real}|/\xi_{real}$). The top row is for Walker, where we outperform the baseline, and the bottom row is for Ball-in-Cup, where we inconsistently outperform the baseline. We observe that learning system parameters accurately is generally important for good policy performance.}
\label{fig:spe}
\end{figure}

We found a strong correlation between how well our model identified the correct simulation parameters and our method's performance improvement over the baseline. Figure \ref{fig:spe} shows a few representative simulation parameter error curves which illustrate that environments in which our agent outperforms the baseline are typically the environments where simulation parameters earned accurately. The agent's performance is most sensitive to the dynamics parameters, which can prove problematic as some dynamics parameters are difficult to learn from sequences of observations and actions.

\subsection{Sim-to-Real Transfer}
\begin{figure*}[h!]
\centering
\begin{subfigure}[t]{0.2\textwidth}
    \centering
    \includegraphics[height=1in]{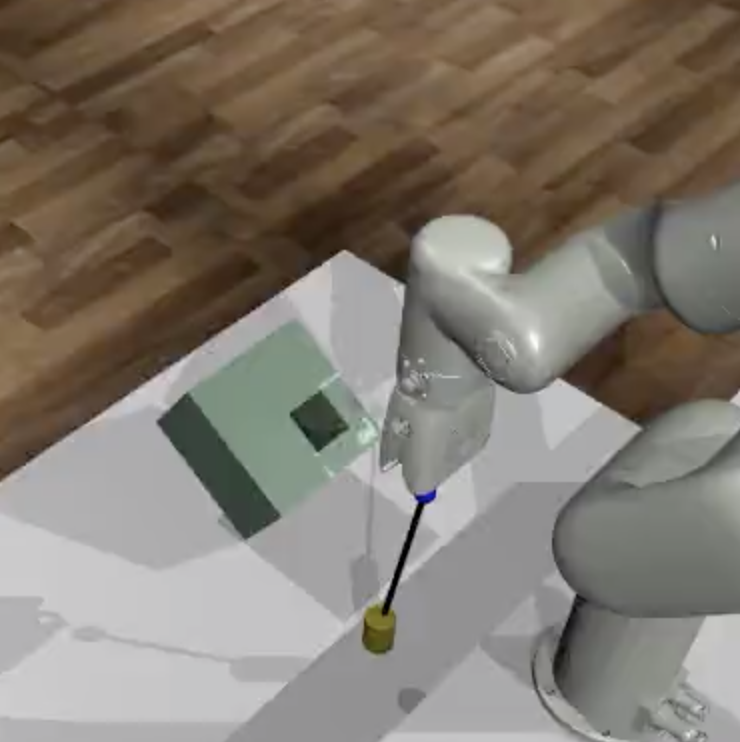}
    \caption{Training Env}
\end{subfigure}%
~
\begin{subfigure}[t]{0.2\textwidth}
    \centering
    \includegraphics[height=1in]{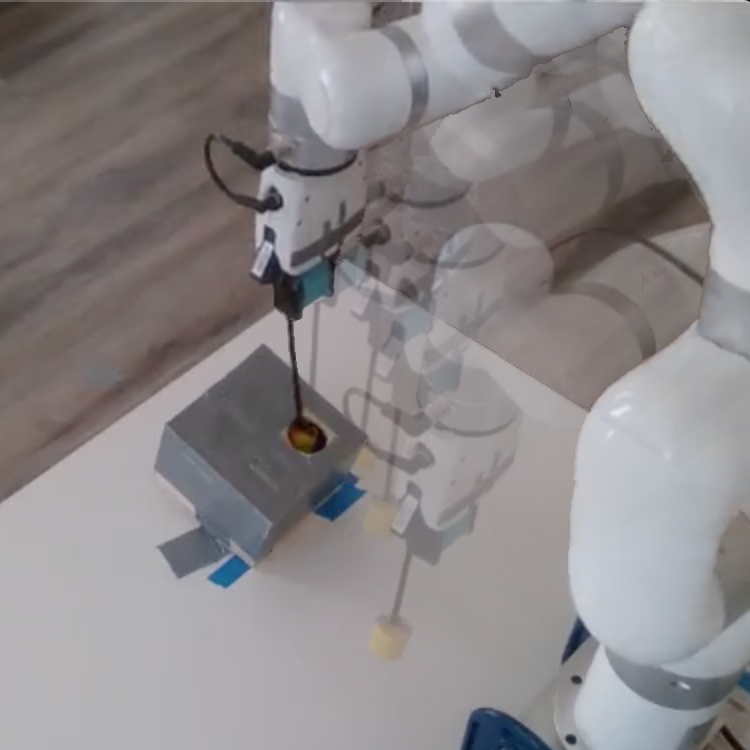}
    \caption{Successful Traj -- Ours}
\end{subfigure}%
~
\begin{subfigure}[t]{0.2\textwidth}
    \centering
    \includegraphics[height=1in]{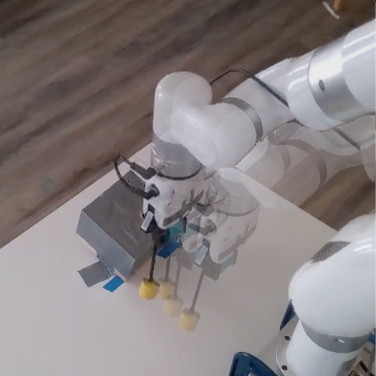}
    \caption{Avg Traj -- DR}
\end{subfigure}%
~
\begin{subfigure}[t]{0.2\textwidth}
    \centering
    \includegraphics[height=1in]{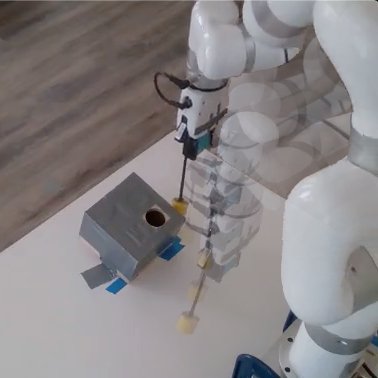}
    \caption{Avg Traj -- Ours}
\end{subfigure}%
\vspace{-0.05in}
\caption{Rope Peg-in-Hole task. The policy transferred from the DR baseline generally does not move towards the hole, whereas our method, trained on $<20\%$ of the simulation policy steps used in the baseline, leads to a policy that consistently moves towards the hole and occasionally succeeds.}
\label{fig:s2r_rope}

\vspace{0.08in}
\begin{subfigure}[t]{0.2\textwidth}
    \centering
    \includegraphics[height=1in]{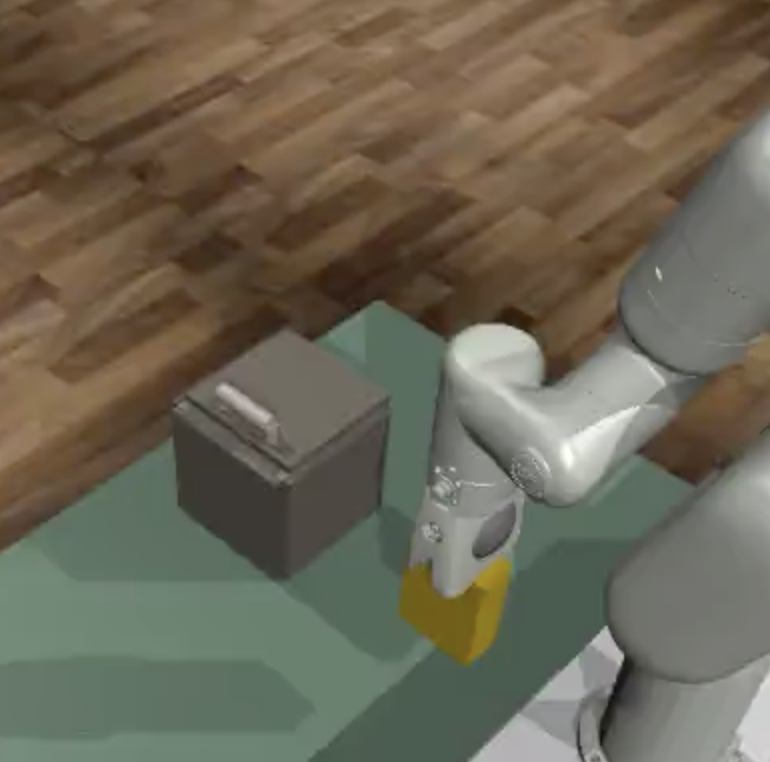}
    \caption{Training Env}
\end{subfigure}%
~
\begin{subfigure}[t]{0.2\textwidth}
    \centering
    \includegraphics[height=1in]{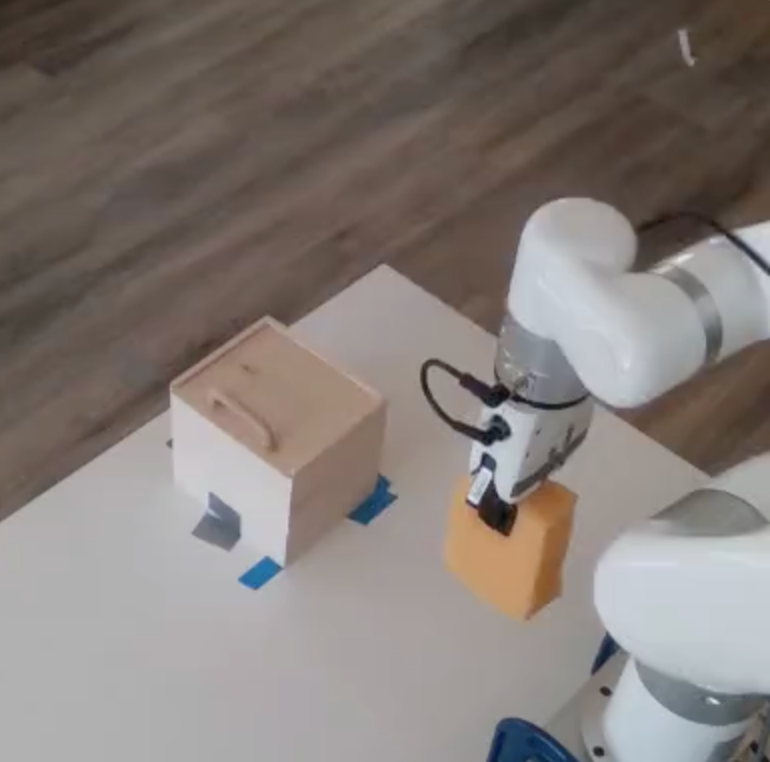}
    \caption{Test Env}
\end{subfigure}%
~
\begin{subfigure}[t]{0.2\textwidth}
    \centering
    \includegraphics[height=1in]{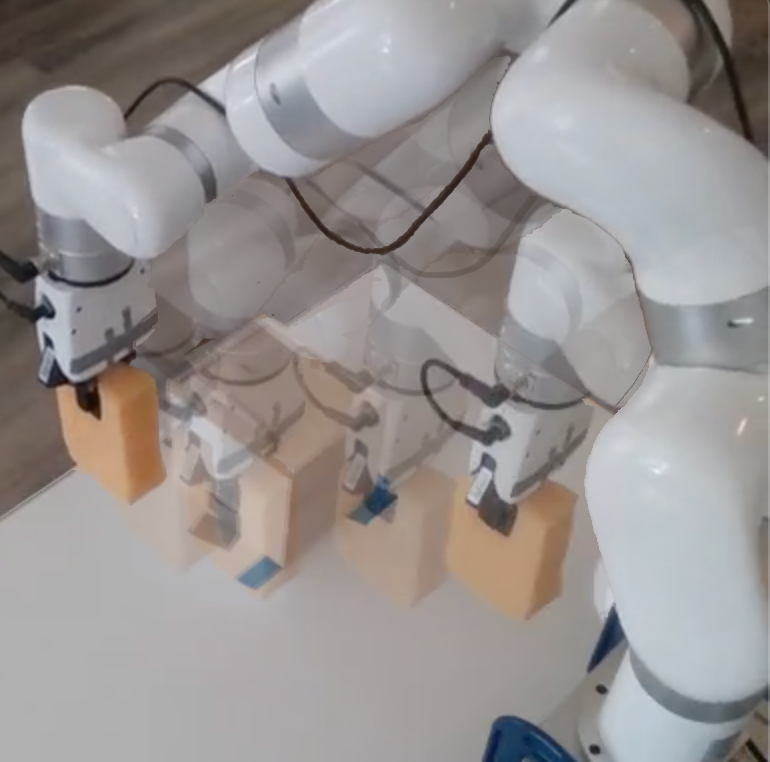}
    \caption{Avg Traj -- DR}
\end{subfigure}%
~
\begin{subfigure}[t]{0.2\textwidth}
    \centering
    \includegraphics[height=1in]{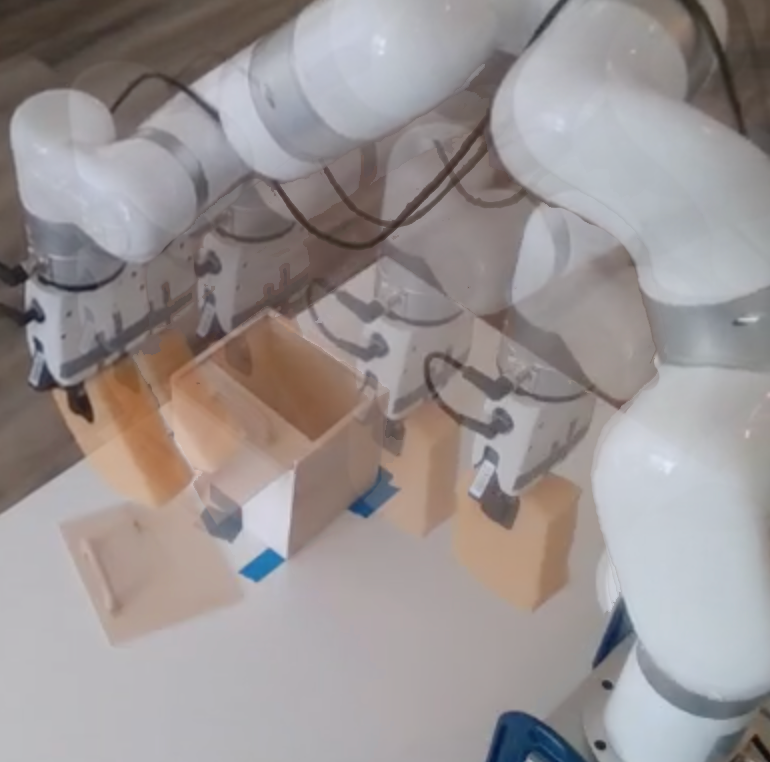}
    \caption{Avg Traj -- Ours}
\end{subfigure}
\vspace{-0.05in}
\caption{Cabinet Slide task. Transferring the DR baseline leads to a policy that moves towards the handle but overshoots to its left, whereas our method, trained on $<20\%$ of the simulation policy steps used in the baseline, leads to a policy that moves to the right of the handle and consistently succeeds at sliding open the cabinet lid.}
\label{fig:s2r_cab}
\vspace{-0.05in}
\end{figure*}

We evaluate our method on sim-to-real transfer on the same two robotic arm tasks (Rope Peg-in-Hole and Cabinet Slide) which we used for sim-to-sim transfer. We implement a real world setup as shown in Figures \ref{fig:s2r_rope} and \ref{fig:s2r_cab}. We use the UFactory xArm 7 and capture RGB observations using a Realsense camera placed at an over-the-shoulder angle.

As with our sim-to-sim experiments, we initialize the training distribution with a more challenging `misparametrized' distribution with  $\frac{3}{4}$x or $\frac{4}{3}$x of the parameters from our simulated pseudo-real world for the Rope Peg-in-Hole task and with $\frac{3}{2}$x or $\frac{2}{3}$x for the Cabinet Slide task. For sample efficiency, we pretrain both the policy and the SPM using domain randomization in the initial misparametrized training environments, then fine tune the policy and simulators jointly using our method with real robot rollouts until the task is successfully completed in the real world. We compare against a domain randomization baseline policy trained to convergence in the initial misparametrized training environment, which also required 5x more additional policy steps than our method. To evaluate the approaches, we look at ten rollouts from each method. Success rates are:
\begin{table}[h!]
\vspace{-0.1in}
\centering
\begin{tabular}{c | c | c }
     Environment & Ours - Success & DR - Success  \\
     \hline
     Rope Peg-in-Hole & 40\% & 0\% \\
     Cabinet Slide & 90\% & 0\%
\end{tabular}
\label{tab:sim2real}
\vspace{-10pt}
\end{table}

\noindent For the rope task, only using RGB information made it challenging to learn a successful policy (e.g. trials may look successful from the camera angle but were not truly successful). As such, future work can incorporate a depth sensor. Due to this challenge our method was only successful occasionally, but qualitatively comparing trajectories in Figure \ref{fig:s2r_rope} we see that the baseline does not move the peg towards the hole, whereas our method consistently moves towards the hole and occasionally succeeds at insertion. In the table, we redefine ``success'' to include near-success, where the robot moves the object over the hole, whether or not it actually enters. For the cabinet task, as shown in Figure \ref{fig:s2r_cab}, our method learns to consistently succeed, whereas the baseline moves towards the box but consistently overshoots and fails.

\section{Conclusion and Discussion}
\label{sec:conclusion}

We proposed an approach to automatically bring the distribution of system parameters in simulation close to that in the real-world through an iterative search process.
In our current work, we only randomize a small set of dynamics parameters, and we intentionally chose parameters with non-interacting effects on the dynamics of the scene. This is sufficient for some tasks, as we showed by transferring policies to the real world, but may not be enough for tasks which require more dexterous manipulation. Future work can apply our method to more complex tasks where many parameters must be learned simultaneously.

The main limitation of our method is that it is difficult to learn the true simulation parameters when the initial randomization is too far from the true values. While our model sometimes converges to the correct parameters even when the true value is outside the initial distribution range, learning is more challenging in this setting.  Choosing an appropriate initial randomization range may require some tuning, but this process should be simpler than individually tuning each  parameter as required for domain randomization.

Our method is most helpful for tasks like object manipulation which depend on particular visual and dynamics parameters which are challenging to measure directly. Our approach is unlikely to be helpful in environments which are very sensitive to particular hard-to-model system parameters, such as the trajectory of a leaf blowing through the wind. We also do not recommend our method for tasks which are not sensitive to system dynamics, such as navigation. While our method could still provide realistic visual randomization, domain randomization is often sufficient in these cases. While domain randomization is a strong baseline, we have shown successful transfers to simulation and reality in cases where naive domain randomization performs poorly. Our method demonstrates that using just unsupervised real-world videos we are able to adjust the randomization mean to be closer to the real system parameters, ultimately leading to improved transfer success in the real-world.

\begin{figure}[]
\centering
    \begin{subfigure}[t]{0.25\linewidth}
    \centering
    \includegraphics[height=.6in]{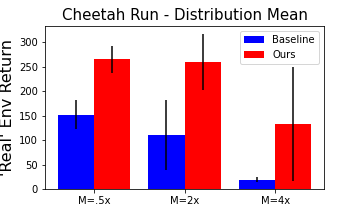}
\end{subfigure}%
~
    \begin{subfigure}[t]{0.25\linewidth}
    \centering
    \includegraphics[height=.6in]{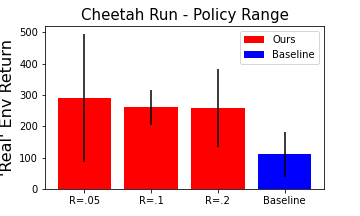}
\end{subfigure}%
~
      \begin{subfigure}[t]{0.25\linewidth}
    \centering
    \includegraphics[height=.6in]{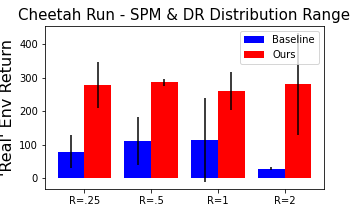}
\end{subfigure}%
\\
\begin{subfigure}[t]{0.25\linewidth}
    \centering
    \includegraphics[height=.6in]{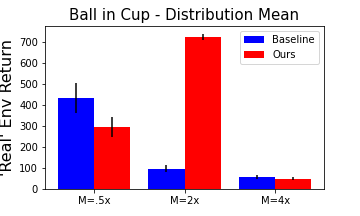}
\end{subfigure}%
~
   \begin{subfigure}[t]{0.25\linewidth}
    \centering
    \includegraphics[height=.6in]{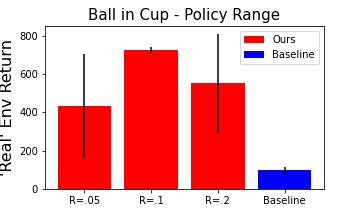}
\end{subfigure}%
~
\begin{subfigure}[t]{0.25\linewidth}
    \centering
    \includegraphics[height=.6in]{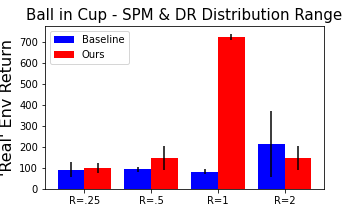}
\end{subfigure}%

\vspace{-.05in}
\caption{Ablations: In envs like Cheetah where the system parameters are learned easily our method is robust to different hyperparameters, but in envs such as Ball in Cup, which are sensitive to hard-to-learn dynamics, our method is less stable. Results are averaged over 3 seeds with std error bars.}
\label{fig:ablation}
\vspace{-.1in}
\end{figure}

\section*{Acknowledgements}
We thank Albert Zhan \& Philip Zhao for providing the xArm experimental environment and code-base. The work is supported by Berkeley Deep Drive and ONR PECASE N00014161723. YD is supported by Microsoft, BAIR Fellowship and OW by DARPA XAI and NSF Fellowship. DP is supported by Google FRA and DARPA Machine Common Sense grant.

\bibliographystyle{IEEEtran}
\bibliography{IEEEabrv,main}

\begin{thebibliography}{10}
\providecommand{\url}[1]{#1}
\csname url@samestyle\endcsname
\providecommand{\newblock}{\relax}
\providecommand{\bibinfo}[2]{#2}
\providecommand{\BIBentrySTDinterwordspacing}{\spaceskip=0pt\relax}
\providecommand{\BIBentryALTinterwordstretchfactor}{4}
\providecommand{\BIBentryALTinterwordspacing}{\spaceskip=\fontdimen2\font plus
\BIBentryALTinterwordstretchfactor\fontdimen3\font minus
  \fontdimen4\font\relax}
\providecommand{\BIBforeignlanguage}[2]{{%
\expandafter\ifx\csname l@#1\endcsname\relax
\typeout{** WARNING: IEEEtran.bst: No hyphenation pattern has been}%
\typeout{** loaded for the language `#1'. Using the pattern for}%
\typeout{** the default language instead.}%
\else
\language=\csname l@#1\endcsname
\fi
#2}}
\providecommand{\BIBdecl}{\relax}
\BIBdecl

\bibitem{tan2018simtoreal}
J.~Tan, T.~Zhang, E.~Coumans, A.~Iscen, Y.~Bai, D.~Hafner, S.~Bohez, and
  V.~Vanhoucke, ``Sim-to-real: Learning agile locomotion for quadruped
  robots,'' 2018.

\bibitem{lee2020learning}
J.~Lee, J.~Hwangbo, L.~Wellhausen, V.~Koltun, and M.~Hutter, ``Learning
  quadrupedal locomotion over challenging terrain,'' \emph{Science Robotics},
  2020.

\bibitem{openai2018learning}
OpenAI, M.~Andrychowicz, B.~Baker, M.~Chociej, R.~Jozefowicz, B.~McGrew,
  J.~Pachocki, A.~Petron, M.~Plappert, G.~Powell, A.~Ray, J.~Schneider,
  S.~Sidor, J.~Tobin, P.~Welinder, L.~Weng, and W.~Zaremba, ``Learning
  dexterous in-hand manipulation,'' 2018.

\bibitem{Jacobi:1995:NRG:645300.648380}
N.~Jacobi, P.~Husbands, and I.~Harvey, ``Noise and the reality gap: The use of
  simulation in evolutionary robotics,'' in \emph{Proceedings of the Third
  European Conference on Advances in Artificial Life}.\hskip 1em plus 0.5em
  minus 0.4em\relax London, UK, UK: Springer-Verlag, 1995, pp. 704--720.

\bibitem{DBLP:journals/corr/LevinePKQ16}
S.~Levine, P.~Pastor, A.~Krizhevsky, and D.~Quillen, ``Learning hand-eye
  coordination for robotic grasping with deep learning and large-scale data
  collection,'' \emph{CoRR}, vol. abs/1603.02199, 2016.

\bibitem{Ljung:1986:SIT:21413}
L.~Ljung, \emph{System Identification: Theory for the User}.\hskip 1em plus
  0.5em minus 0.4em\relax Upper Saddle River, NJ, USA: Prentice-Hall, Inc.,
  1986.

\bibitem{ganin2015domainadversarial}
Y.~Ganin, E.~Ustinova, H.~Ajakan, P.~Germain, H.~Larochelle, F.~Laviolette,
  M.~Marchand, and V.~Lempitsky, ``Domain-adversarial training of neural
  networks,'' 2015.

\bibitem{tobin2017domain}
J.~Tobin, R.~Fong, A.~Ray, J.~Schneider, W.~Zaremba, and P.~Abbeel, ``Domain
  randomization for transferring deep neural networks from simulation to the
  real world,'' in \emph{2017 IEEE/RSJ International Conference on Intelligent
  Robots and Systems (IROS)}.\hskip 1em plus 0.5em minus 0.4em\relax IEEE,
  2017, pp. 23--30.

\bibitem{peng2018sim}
X.~B. Peng, M.~Andrychowicz, W.~Zaremba, and P.~Abbeel, ``Sim-to-real transfer
  of robotic control with dynamics randomization,'' in \emph{2018 IEEE
  international conference on robotics and automation (ICRA)}.\hskip 1em plus
  0.5em minus 0.4em\relax IEEE, 2018, pp. 1--8.

\bibitem{DBLP:journals/corr/abs-1806-07851}
J.~Matas, S.~James, and A.~J. Davison, ``Sim-to-real reinforcement learning for
  deformable object manipulation,'' \emph{CoRR}, vol. abs/1806.07851, 2018.

\bibitem{simopt}
Y.~Chebotar, A.~Handa, V.~Makoviychuk, M.~Macklin, J.~Issac, N.~D. Ratliff, and
  D.~Fox, ``Closing the sim-to-real loop: Adapting simulation randomization
  with real world experience,'' \emph{CoRR}, vol. abs/1810.05687, 2018.

\bibitem{chang2020sim2real2sim}
P.~Chang and T.~Padir, ``Sim2real2sim: Bridging the gap between simulation and
  real-world in flexible object manipulation,'' \emph{arXiv preprint
  arXiv:2002.02538}, 2020.

\bibitem{pinto2017asymmetric}
L.~Pinto, M.~Andrychowicz, P.~Welinder, W.~Zaremba, and P.~Abbeel, ``Asymmetric
  actor critic for image-based robot learning,'' \emph{arXiv preprint
  arXiv:1710.06542}, 2017.

\bibitem{CAD2RL}
F.~Sadeghi and S.~Levine, ``(cad){\textdollar}{\^{}}2{\textdollar}rl: Real
  single-image flight without a single real image,'' \emph{CoRR}, vol.
  abs/1611.04201, 2016.

\bibitem{adr}
OpenAI, I.~Akkaya, M.~Andrychowicz, M.~Chociej, M.~Litwin, B.~McGrew,
  A.~Petron, A.~Paino, M.~Plappert, G.~Powell, R.~Ribas, J.~Schneider,
  N.~Tezak, J.~Tworek, P.~Welinder, L.~Weng, Q.~Yuan, W.~Zaremba, and L.~Zhang,
  ``Solving rubik's cube with a robot hand,'' 2019.

\bibitem{RoboImitationPeng20}
X.~B. Peng, E.~Coumans, T.~Zhang, T.-W.~E. Lee, J.~Tan, and S.~Levine,
  ``Learning agile robotic locomotion skills by imitating animals,'' in
  \emph{Robotics: Science and Systems}, 07 2020.

\bibitem{yu2017preparing}
W.~Yu, J.~Tan, C.~K. Liu, and G.~Turk, ``Preparing for the unknown: Learning a
  universal policy with online system identification,'' \emph{arXiv preprint
  arXiv:1702.02453}, 2017.

\bibitem{zhu2017fast}
S.~Zhu, A.~Kimmel, K.~E. Bekris, and A.~Boularias, ``Fast model identification
  via physics engines for data-efficient policy search,'' \emph{arXiv preprint
  arXiv:1710.08893}, 2017.

\bibitem{jeong2019selfsupervised}
R.~Jeong, Y.~Aytar, D.~Khosid, Y.~Zhou, J.~Kay, T.~Lampe, K.~Bousmalis, and
  F.~Nori, ``Self-supervised sim-to-real adaptation for visual robotic
  manipulation,'' 2019.

\bibitem{haarnoja2018soft}
T.~Haarnoja, A.~Zhou, P.~Abbeel, and S.~Levine, ``Soft actor-critic: Off-policy
  maximum entropy deep reinforcement learning with a stochastic actor,'' 2018.

\bibitem{laskin_srinivas2020curl}
M.~Laskin, A.~Srinivas, and P.~Abbeel, ``Curl: Contrastive unsupervised
  representations for reinforcement learning,'' \emph{Proceedings of the 37th
  International Conference on Machine Learning, Vienna, Austria, PMLR 119},
  2020, arXiv:2004.04136.

\bibitem{mildenhall2020nerf}
B.~Mildenhall, P.~P. Srinivasan, M.~Tancik, J.~T. Barron, R.~Ramamoorthi, and
  R.~Ng, ``Nerf: Representing scenes as neural radiance fields for view
  synthesis,'' \emph{arXiv preprint arXiv:2003.08934}, 2020.

\bibitem{deepmindcontrolsuite2018}
Y.~Tassa, Y.~Doron, A.~Muldal, T.~Erez, Y.~Li, D.~de~Las~Casas, D.~Budden,
  A.~Abdolmaleki, J.~Merel, A.~Lefrancq, T.~Lillicrap, and M.~Riedmiller,
  ``Deep{Mind} control suite,'' DeepMind, Tech. Rep., Jan. 2018.

\bibitem{RAD}
M.~Laskin, K.~Lee, A.~Stooke, L.~Pinto, P.~Abbeel, and A.~Srinivas,
  ``Reinforcement learning with augmented data,'' \emph{arXiv preprint
  arXiv:2004.14990}, 2020.

\bibitem{lin2020nerfpytorch}
L.~Yen-Chen, ``Nerf-pytorch,''
  \url{https://github.com/yenchenlin/nerf-pytorch/}, 2020.

\end{thebibliography}

\begin{appendices}

\section{Model Architecture}\label{appendix:architecture}
For all our experiments we use the same encoder and actor-critic architecture as found in Appendix E of \cite{RAD}. We also use the same hyperparameters from \cite{RAD}, with the following deviations: we use batch size 128, the crop augmentation, and episode lengths of 200 policy steps for DM control tasks and 60 for the robotics tasks. We use 1 million steps for the DM control environments and 500k steps for the robotics tasks.

The Search Param Model is given the set of system parameters and a trajectory composed of a sequence of images and the corresponding actions in between. To embed the system parameters, we use the positional encoding from \cite{mildenhall2020nerf}, as implemented in \cite{lin2020nerfpytorch}. To embed the images, we use the same encoder architecture as in \cite{RAD}. We concatenate together the embedded parameters, encoded image features, and actions, which are then passed through a 2 layer MLP with 400 hidden units each, with an ELU activation and a dropout layer ($p=.5$). We train the model using logistic regression, optimizing with Adam with learning rate $= 1e^{-3}$ and $\beta=0.9$.

\end{appendices}

\end{document}